\documentclass[11pt,a4paper]{article}
\usepackage[hyperref]{naaclhlt2019}
\usepackage{times}
\usepackage{latexsym}
\usepackage{graphicx}
\usepackage{url}
\usepackage{fixltx2e}

\usepackage{array,multirow,graphicx}
\usepackage{float}

\aclfinalcopy 
\def\aclpaperid{4083} 

\title{Robust Semantic Parsing with Adversarial Learning for Domain Generalization}

\author{Gabriel Marzinotto$^{1,2}$, G\'eraldine Damnati$^{1}$, Fr\'ed\'eric B\'echet$^{2}$, Beno\^it Favre$^{2}$\\
  (1) Orange Labs / Lannion France \\
  (2) Aix Marseille Univ, CNRS, LIS / Marseille France \\
  \{\tt gabriel.marzinotto@orange.com, geraldine.damnati@orange.com \\ \tt frederic.bechet@lis-lab.fr, benoit.favre@lis-lab.fr\} \\}

\date{}

\begin{document}
\maketitle
\begin{abstract}
This paper addresses the issue of generalization for Semantic Parsing in an adversarial framework. Building models that are more robust to inter-document variability is crucial for the integration of Semantic Parsing technologies in real applications. The underlying question throughout this study is whether adversarial learning can be used to train models on a higher level of abstraction in order to increase their robustness to lexical and stylistic variations.
We propose to perform Semantic Parsing with a domain classification adversarial task without explicit knowledge of the domain. The strategy is first evaluated on a French corpus of encyclopedic documents, annotated with FrameNet, in an information retrieval perspective, then on PropBank Semantic Role Labeling task on the CoNLL-2005 benchmark. We show that adversarial learning increases all models generalization capabilities both on in and out-of-domain data. 



\end{abstract}

\section{Introduction}

For many NLP applications, models that perform well on multiple domains and data sources are essential. As data labeling is expensive and time consuming, especially when it requires specific expertise (FrameNet, Universal Dependencies, etc.), annotations for every domain and data source are not feasible. On the other hand, domain biases are a major problem in almost every supervised NLP task. Models learn these biases as useful information and experience a significant performance drop whenever they are applied on data from a different source or domain. A recent approach attempting to tackle domain biases and build robust systems consists in using neural networks and adversarial learning to build domain independent representations. In the NLP community, this method has been mostly used in crosslingual models to transfer information from English to low resource languages in problems and recently in various monolingual tasks in order to alleviate domain bias of trained models.

In the context of Semantic Frame Parsing, we address in this paper the generalization issue of models trained on one or several domains and applied to a new domain.
We show that adversarial learning can be used to improve the generalization capacities of semantic parsing models to out of domain data. We propose an adversarial framework based on a domain classification task that we use as a regularization technique on state-of-the-art semantic parsing systems. We use unsupervised domain inference to obtain labels for the classification task.

Firstly we perform experiments on a large multi-domain frame corpus~\cite{marzinotto:calor} where only a relatively small number of frames where annotated, corresponding to possible targets in an Information Extraction applicative framework. We evaluate our adversarial framework with a semantic frame parser we developed on this corpus and presented in \cite{marzinotto:hal-01731385}.
Secondly we checked the genericity of our approach on the standard PropBank Semantic Role Labeling task on the CoNLL-2005 benchmark, with a tagging model proposed by~\cite{he2017deep}. We show that in both cases adversarial learning increases all models generalization capabilities both on in and out-of-domain data. 





\section{Related Work}
\label{sec:related}

\subsection{Domain-Adversarial Training}

Domain Independence can be approached from different perspectives. A popular approach that emerged in image processing \cite{JMLR:v17:15-239} consists in optimizing a double objective composed of a task-specific classifier and an adversarial domain classifier. The latter is called adversarial because it is connected to the task-specific classifier using a gradient reversal layer. During training a saddle point is searched where the task-specific classifier is good and the domain classifier is bad. It has been shown in \cite{ganin2015unsupervised} that this implicitly optimizes the hidden representation towards domain independence.

In NLP problems this approach has successfully been used to train cross-lingual word representations \cite{conneau2017word} and to transfer learning from English to low resource languages for POS tagging \cite{D17-1302} and sentiment analysis \cite{2016arXiv160601614C}. These approaches introduce language classifiers with an adversarial objective to train task-specific but language agnostic representations. Besides the cross-lingual transfer problem, there are few studies of the impact of domain-adversarial training in a monolingual setup. For instance,  \cite{DBLP:journals/corr/LiuQH17} successfully uses this technique to improve generalization in a document classification task. It has also been used recently for varied tasks such as transfer learning on Q\&A systems \cite{yu2018modelling} or duplicate question detection \cite{Shah2018AdversarialDA} and removal of protected attributes from social media textual data \cite{Elazar2018AdversarialRO}.


\subsection{Robustness in Semantic Frame Parsing}
In Frame Semantic Parsing, data is scarce and classic evaluation settings seldom propose out-of-domain test data. Despite the existence of out-of-domain corpora such MASC \cite{passonneau2012masc} and YAGS \cite{hartmann2017out} the domain adaptation problem has been widely reported \cite{nuges2009,sogaard2015using} but not extensively studied. 
Recently, \cite{hartmann2017out} presented the first in depth study of the domain adaptation problem using the YAGS frame corpus. They show that the main problem in domain adaptation for frame semantic parsing is the frame identification step and propose a more robust classifier using predicate and context embeddings to perform frame identification.
This approach is suitable for cascade systems such as SEMAFOR \cite{das2014frame}, \cite{hermann2014semantic} and \cite{SoinFrameParsing}. In this paper we propose to study the generalization issue within the framework of a sequence tagging semantic frame parser that performs frame selection and argument classification in one step. 


\section{Semantic parsing model with an adversarial training scheme}
\label{sec:models}



\subsection{Semantic parsing model: \texttt{biGRU} }
\label{sec:bigru}
We use in this study a sequence tagging semantic frame parser that performs frame selection and argument classification in one step based on a deep bi-directional GRU tagger ($biGRU$). The advantage of this architecture is its flexibility as it can be applied to several semantic parsing schemes such as PropBank \cite{he2017deep} and FrameNet \cite{SoinFrameParsing}.

More precisely, the model consists of a 4 layer bi-directional Gated Recurrent Unit (GRU) with highway connections \cite{SrivastavaGS15}.
This model does not rely solely on word embeddings as input. Instead, it has a richer set of features including syntactic, morphological and surface features.
(see \cite{marzinotto:hal-01731385} for more details).

Except for words where we use pre-trained embeddings, we use randomly initialized embedding layers for categorical features.


\subsection{Sequence encoding/decoding}
\label{sec:decoding}
For all experiments we use a BIO label encoding.
To ensure that output sequences respect the BIO constrains we implement an A$^*$ decoding strategy similar to the one proposed by \cite{he2017deep}. We further apply a coherence filter to the output of the tagging process. This filter ensures that the predicted semantic structure is acceptable. Given a sentence and a word $w$ that is a Lexical Unit (LU) trigger, we select the frame $F$ as being the most probable frame among the ones that can have $w$ as a trigger. Once $F$ is determined, we then mask all FEs that do not belong to $F$ and perform constrained A$^*$ decoding.
Finally, we improve this strategy by introducing a parameter $\delta \in (-1;1)$ that is added to the output probability of the \textit{null} label $P(y_t=O)$ at each time-step. By default, with $\delta=0$ the most probable non-null hypothesis is selected if its probability is higher than $P(y_t=O)$. Varying $\delta>0$ (resp. $\delta<0$) is equivalent to being more strict (resp. less strict) on the highest non-null hypothesis. By doing so we can study the precision/recall (P/R) trade-off of our models. This $\delta$ parameter is tuned on a validation set and we either provide the P/R curve or report scores for the $Fmax$ setting.  


\subsection{Adversarial Domain Classifier}
\label{sec:adv_model}

In order to design an efficient adversarial task, several criteria have to be met. The task has to be related to the biases it is supposed to alleviate. And furthermore, the adversarial task should not be correlated to the main task (i.e semantic parsing here), otherwise it may harm the model's performances. Determining where these biases lay is not easy, although this is critical for the success of our method. We propose to use a domain classification adversarial task.

Given two data-sets \texttt{(X1, Y1)} and \texttt{(X2, Y2)} from different domains. The expected gains from introducing an adversarial domain classifier are proportional to how different \texttt{X1} and \texttt{X2} are (the more dissimilar the better) and proportional to how similar the label distributions \texttt{Y1} and \texttt{Y2} are (higher similarity is better). Otherwise, if \texttt{X1} and \texttt{X2} are very similar, there is no need to transfer learning from one domain to another. Under this condition, The adversarial domain classifier will not be able to recognize domains and give proper information on how to build better representations. If \texttt{Y1} and \texttt{Y2} are extremely dissimilar, to the point where \texttt{Y} cannot be predicted without explicit domain information, using adversarial learning may be harmful. In our case, prior probabilities for both frame distribution and word senses are correlated to the thematic domains. However, adversarial learning can still be useful because most of the LUs are polysemous even within a domain. Therefore, the model needs to learn word sense disambiguation through a more complex process than simply using the domain information.

Our adversarial domain classifier is an extension of \cite{ganin2014unsupervised} to recurrent neural networks. We start from our $biGRU$ semantic parser and on the last hidden layer, we stack another neural network that implements a domain classifier (called adversarial task). The domain classifier is connected to the $biGRU$ using a gradient reversal layer. Training consists in finding a saddle point where the semantic parser is good and the domain classifier is bad. This way, the model is optimized to be domain independent.

\begin{figure}[htbp]
  \centering
  \includegraphics[width=1\linewidth]{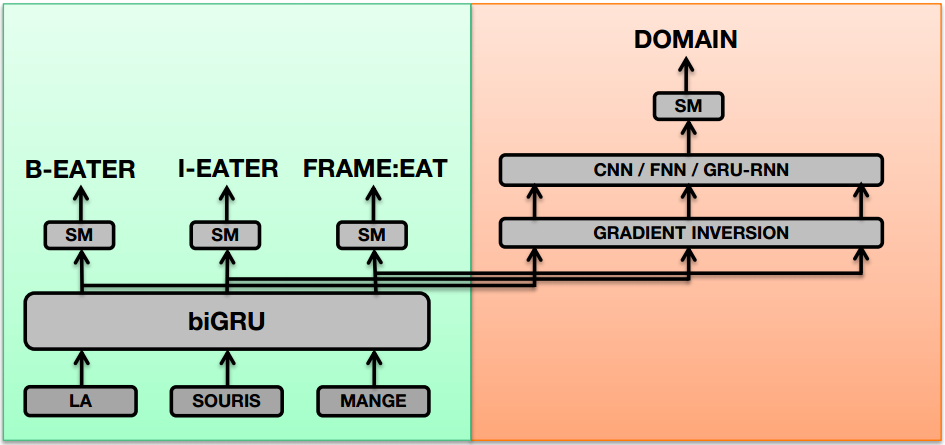}
  \caption{Adversarial Domain Classifier model}
  \label{fig:al1}
\end{figure}

The architecture is shown in Figure \ref{fig:al1}. The adversarial task can be implemented using a CNN, a RNN or a FNN. In this paper we use CNN as they yield the best results on preliminary experiments.

This architecture is trained following the guidelines of \cite{ganin2014unsupervised}. During training, the adversarial classifier is trained to predict the target class (\textit{i.e.} to minimize the loss $L_{adv}$) while the main task model is trained to make the adversarial task fail (\textit{i.e.} to minimize the loss $L_{frame}-L_{adv}$). In practice, in order to ensure stability when the parameters of the shared network are updated, the adversarial task gradient magnitude is attenuated by a factor $\lambda$ as shown in (\ref{eq:1}). Here $\nabla L$ represents the gradients w.r.t the weights $\theta$ for either the frame classifier loss or the adversarial loss, $\theta$ are the model's parameters being updated, and $\mu$ is the learning rate. This $\lambda$ factor increases on every epoch following (\ref{eq:2}), where $p$ is the progress, starting at 0 and increasing linearly up to 1 at the last epoch.  
\begin{equation} \label{eq:1}
 \theta \leftarrow \theta - \mu * (\nabla L_{frame} - \lambda  \nabla L_{adv})
\end{equation}
\begin{equation} \label{eq:2}
 \lambda = \frac{2}{1 + \exp(- 10 \cdot p)} -1 
\end{equation}


\subsection{Unsupervised domain inference}
\label{sec:domain-inference}
The originality of our approach lies in the design of an adversarial task in an unsupervised way. Our purpose is to design a general method that could easily and efficiently apply in any realistic conditions, independently of the fact that the training sentences can be linked to an \textit{a priori} explicit domain or topic.
To this end, an unsupervised clustering algorithm is used to partition the training corpus into \textit{clusters} that are supposed to reflect topics, lexical or stylistic variation within the corpus, depending on the metric used for the clustering. In a first attempt for our experiments, we use K-means, in the \texttt{sklearn} implementation, to cluster training sentences. For clustering purpose, sentences are represented by the average of their word embedding vectors. K-means with a euclidean distance is then used to group these representations into $K$ clusters. We use a standard \textit{Kmeans++} initialization. The clustering process is repeated 10 times and the one that produces the minimal intra-cluster inertia is kept.
Eventually, the resulting clusters are used as classes that the CNN will be trained to recognize for each corresponding training sentence. The underlying assumption is that the clustering process will capture domain-related regularities and biases that will be harnessed by the adversarial task in order to increase the model's generalization capacities.

\section{Evaluation setting}
\label{sec:corpus}
To create an experimental setting that shows the effect of domain on the semantic parsing task, we run experiments on the CALOR-Frame corpus \cite{marzinotto:calor}, which is a compilation of French encyclopedic documents, with manual FrameNet annotations \cite{baker1998berkeley}. 
 The CALOR-Frame corpus has been designed in the perspective of Information Extraction tasks (automatic filling of Knowledge Bases, document question answering, semantically driven reading comprehension, etc.). Due to the \textit{partial parsing} policy, the CALOR corpus presents a much higher amount of occurrences per Frame (504) than the FrameNet 1.5 corpus (33). 


We selected three subsets from the corpus, each one from a different source and/or thematic domain:  Wikipedia World War 1 portal (\texttt{D1}), Wikipedia Archeology portal (\texttt{D2}) and Vikidia\footnote{Vikidia is an encyclopedia for children \url{vikidia.org}} Ancient history portal (\texttt{D3}). These sources allow to study the impact of both style changes (associated to differences on syntactic structures) and thematic changes (associated to lexical differences). 

\begin{table}
\small
\centering
\begin{tabular}{|l|c|c|c|c|}
\hline
\textbf{Document Source}              & \textbf{\# Sentence} & \textbf{\# Frame} & \textbf{\# FE} \\ \hline
\texttt{D1} Wikipedia WW1             & 30994             & 14227              & 32708        \\ \hline
\texttt{D2} Wikipedia ARCH             & 27023                    & 9943               & 19892            \\ \hline
\texttt{D3} Vikidia ANC  & 5841                & 1617                &  3246           \\ \hline
\end{tabular}
\normalsize
\caption{Description of the CALOR-Frame corpus}
\label{tab:corpus}
\end{table}

\begin{figure*}[htbp]
\centering
  \includegraphics[width=0.9\linewidth]{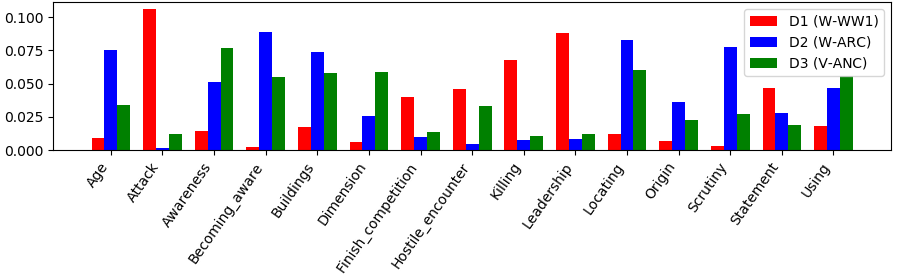}  \caption{Most frequent frames and their normalized distribution for each partition (from left to right: \texttt{W-WW1} (\texttt{D1}), \texttt{W-ARC} (\texttt{D2}) and \texttt{V-ANC} (\texttt{D3}))}
  \label{fig:frame_dist}
\end{figure*}

For the study, we focus on a set of 53 Frames that occur in all the three sub-corpora. Each partition has a different prior on the Frames and lexical units (LUs) distributions. Figure \ref{fig:frame_dist} shows the normalized Frame distributions for the three subsets, illustrating the thematic domain dependence. Frames such as  \textit{Attack}  and \textit{Leadership} are frequent in \texttt{D1} while \textit{Age} and \textit{Scrutiny} are characteristic Frames for \texttt{D2}. The same analysis can be done using LUs, yielding similar conclusions. We also observe that \texttt{D2} and \texttt{D3} are more similar but the difference between these sub-corpora lie more in the syntactic structure of sentences and in the polysemic use of some LU, as will be discussed in the experiments. 


\section{Results}
\label{sec:results}

\begin{figure}[htbp]
  \centering
  \includegraphics[width=1.0\linewidth]{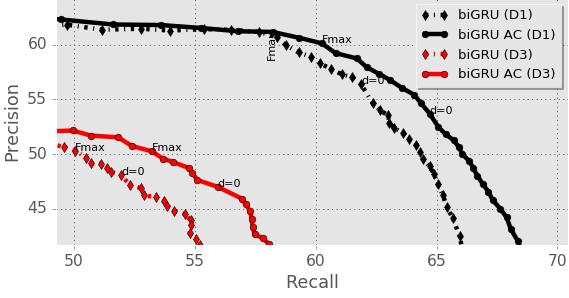}
  \caption{Precision/Recall trade-off with ($biGRU$+$AC$) and without ($biGRU$) adversarial training on $D1$ (in-domain) and $D3$ (out-of-domain)}
  \label{fig:pr_curve}
\end{figure}  


In these experiments we evaluate the impact of our adversarial training scheme on the semantic frame parser presented in section~\ref{sec:bigru}. The parser is trained on 80\% of the $D1$ and $D2$ corpora and evaluated on the remaining 20\%, considered here as \textit{in-domain} data, as well as the whole $D3$ corpus for the \textit{out-of-domain} data.
For the domain inference, we use the method presented in section~\ref{sec:domain-inference}. Several experiments were made, varying the number of clusters $K$ ($K=2$, $K=5$ and $K=10$). The performance obtained were very similar, the influence of $K$ being negligible in our experiments, therefore we report here only those done with $K=5$. 

We report results using \textit{precision/recall/f-measure} metrics at the \textit{Target}, \textit{Frame} and \textit{Argument identification} levels. The errors are cumulative: to obtain a correct argument identification, we need to have both its frame and target correct. Moreover we use a hard-span metric for argument identification, meaning that both the label and the span have to be correct for an hypothesis to be considered as correct.

Results are given in figure~\ref{fig:pr_curve} where the precision/recall curve on argument identification is given with and without adversarial training for the in-domain corpus $D1$ and the out-of domain corpus $D3$. Table~\ref{tab:frame_arch_hist} presents F-measure (F-max) for the 3 identification tasks (target, frame, argument) for $D1$, $D2$ and $D3$.





Figure \ref{fig:pr_curve} clearly illustrates the drop in performance between in-domain and out-of-domain corpora. The difference is significant, it accumulates at each step resulting on a 9 points drop in F-max for the argument identification task as shown in table~\ref{tab:frame_arch_hist}.
When applying our adversarial method during model training, we clearly increase the generalization capabilities of our model on out-of-domain data ($D3$), as the $biGRU+AC$ curve outperforms the $biGRU$ curve at every operating point in figure~\ref{fig:pr_curve}. This is confirmed on the F-max values for each level in table~\ref{tab:frame_arch_hist}.

Interestingly our adversarial training method not only improves parser performance on out-of-domain data, but also on in-domain data, as shown for $D1$ in figure~\ref{fig:pr_curve}. This improvement is mainly due to a gain in recall, and is confirmed on the F-max measures for $D1$ and $D2$ in table~\ref{tab:frame_arch_hist}. 

We believe that, since domains are correlated with the lexicon, the adversarial objective pushes the model to rely less on lexical features and to give more importance to features such as syntax and part-of-speech. This is inline with the observation that most of the improvement comes from a higher recall. This can also explain the performance gains on in-domain data. The high dimensionality of word embeddings may have lead to some over-fitting on our initial $biGRU$ model. On the other hand, adversarial learning can act as a regularization technique that makes the model rely as much as possible on domain independent features.   



\begin{table*}
\centering
\small
\begin{tabular}{@{\extracolsep{2pt}}lccccccccc@{}}
\hline
  &  \multicolumn{3}{c}{Target Identification}  &  \multicolumn{3}{c}{Frame Identification} &  \multicolumn{3}{c}{Argument Identification}\\
 \cline{2-4}  \cline{5-7}  \cline{8-10}
  & \multicolumn{2}{c}{in-domain} & \multicolumn{1}{c}{out-of-domain} &  \multicolumn{2}{c}{in-domain} & \multicolumn{1}{c}{out-of-domain} &  \multicolumn{2}{c}{in-domain} & \multicolumn{1}{c}{out-of-domain} \\
  &  \multicolumn{1}{c}{\texttt{D1}} & \multicolumn{1}{c}{\texttt{D2}} & \multicolumn{1}{c}{\texttt{D3}}  &  \multicolumn{1}{c}{\texttt{D1}} & \multicolumn{1}{c}{\texttt{D2}} & \multicolumn{1}{c}{\texttt{D3}}  &  \multicolumn{1}{c}{\texttt{D1}} & \multicolumn{1}{c}{\texttt{D2}} & \multicolumn{1}{c}{\texttt{D3}} \\ 
 \hline
 \multicolumn{1}{l}{$biGRU$}        & 97.1 & 97.5 & 94.2 & 93.4 & 95.4 & 91.0 & 59.2 & 56.3 & 50.2 \\ 
 \multicolumn{1}{l}{$biGRU$+$AC$}   & 97.3 & \textbf{98.7} & 94.7 & 94.2 & \textbf{95.9} & 91.9 & 60.0 & \textbf{57.0} & \textbf{51.7} \\ \hline
 \multicolumn{1}{l}{$biGRU$+$AC$-$gold$}   & \textbf{97.7} & 98.2 & \textbf{94.9} & \textbf{94.9} & \textbf{95.9} & \textbf{92.0} & \textbf{60.1} & 56.7 & 51.3 \\ 
\hline
\end{tabular}
\normalsize

\caption{F-measure (Fmax) on target, frame and argument identification with ($biGRU$+$AC$) and without ($biGRU$) adversarial training. Clustering with exact domain labels is given in line $biGRU$+$AC$-$gold$}
\label{tab:frame_arch_hist}
\end{table*}

In order to have a better understanding of the behavior of our method we performed two additional contrastive experiments where we used firstly \textit{gold domain labels} instead of inferred ones, and secondly a single domain corpus for training rather than a multi-domain one.


\paragraph{Gold domain labels.} We consider here the \textit{true} domain labels for $D1$ and $D2$ as the classes for our adversarial classifier. Therefore only two domains are considered in the adversarial training process.
Results are presented in the second part of table \ref{tab:frame_arch_hist}, in line $biGRU$+$AC$-$gold$. As we can see results are very similar to those obtained with automatic clustering ($biGRU$+$AC$). With an average difference of only $0.3$pts (Fmax) for the argument identification task across the different domains. This confirms that our unsupervised clustering approach is as efficient as a supervised domain classification for adversarial learning. Moreover, our approach has the advantage that it can be applied in situations where no domain information is available.

\paragraph{Single-thematic corpus:} This contrastive experiment consists in using as training a single-domain corpus. We want to check if the gains obtained on both in and out-of-domain corpora in table~\ref{tab:frame_arch_hist} hold when the training corpus does not explicitly contain several domains. Here, the models are trained only on the training set from \texttt{D1}. We evaluate them on \texttt{D1} (in-domain) and  \texttt{D2,D3} (out-of-domain). 
The adversarial task is obtained by running our domain inference algorithm only on \texttt{D1} training set. Here again, we have chosen to partition training data into 5 clusters.
Alternative experiments not reported in this paper using only \texttt{D2} as in-domain training data have also been performed and yielded similar conclusions.
F-max values reported in table~\ref{tab:frame_hist} are lower than those of table~\ref{tab:frame_arch_hist}. This is expected as the training corpus considered here is much smaller (only $D1$), however performances follow the same trend: some gains are obtained for all 3 levels both for in and out-of-domain corpora. This is a very interesting result as it shows that there is no need for an explicitly multi-thematic training corpus in order to run the adversarial training and to obtain some gains in terms of model generalization.

\begin{table*}
\centering
\small
\begin{tabular}{@{\extracolsep{3pt}}lccccccccc@{}}
\hline
  &  \multicolumn{3}{c}{Target Identification}  &  \multicolumn{3}{c}{Frame Identification} &  \multicolumn{3}{c}{Argument Identification}\\
 \cline{2-4}  \cline{5-7}  \cline{8-10}
  &  \multicolumn{1}{c}{in-domain} & \multicolumn{2}{c}{out-of-domain} &  \multicolumn{1}{c}{in-domain} & \multicolumn{2}{c}{out-of-domain} &  \multicolumn{1}{c}{in-domain} & \multicolumn{2}{c}{out-of-domain} \\
  &  \multicolumn{1}{c}{\texttt{D1}} & \multicolumn{1}{c}{\texttt{D2}} & \multicolumn{1}{c}{\texttt{D3}}  &  \multicolumn{1}{c}{\texttt{D1}} & \multicolumn{1}{c}{\texttt{D2}} & \multicolumn{1}{c}{\texttt{D3}}  &  \multicolumn{1}{c}{\texttt{D1}} & \multicolumn{1}{c}{\texttt{D2}} & \multicolumn{1}{c}{\texttt{D3}} \\ 
 \hline
  \multicolumn{1}{l}{$biGRU$}        & 97.6 & 95.5 & 93.3 & 93.8 & 93.4 & 90.9 & 58.2 & 46.1 & 43.6\\ 
  \multicolumn{1}{l}{$biGRU$+$AC$}   & 97.6 & 95.6 & 94.3 & 95.3 & 94.5 & 91.2 & 60.0 & 47.1 & 45.2 \\ 
 \hline
\end{tabular}
\normalsize
\caption{Frame semantic parsing performances (Fmax). Models trained on \texttt{D1}. Adversarial learning with inferred domains $biGRU+AC$.}
\label{tab:frame_hist}
\end{table*}

\subsection{Error Analysis}

\subsubsection{Target and Frame Identification}

\begin{table}
\centering
\small
\begin{tabular}{lccccc}
\hline
LU & \multicolumn{2}{c}{$biGRU$} & \multicolumn{2}{c}{$biGRU+AC$} \\
  & \texttt{D1} & \texttt{D3} & \texttt{D1} & \texttt{D3} \\
\hline
\texttt{arriver}    & 88.2 & 63.4 & 91.8 & 70.0 \\ 
\texttt{\'{e}crire}    & 96.5 & 73.3 & 97.8 & 88.4 \\ 
\texttt{expression} & 49.8 & 66.6 & 56.5 & 92.4 \\ 
\hline
\end{tabular}
\normalsize
\caption{Frame Identification score for LUs with the highest variation in cluster distribution }
\label{tab:LU_FrameID}
\end{table}

When looking carefully at the generalization capabilities of the initial model, we observed that the frames with the highest performance drops on \texttt{D3} are those associated to LUs that are polysemous in a general context, but are unambiguous given a thematic domain. 
For example, \textit{installer (to install)} triggers the frame \textit{Installing} in both \texttt{D1} and \texttt{D2}, but triggers \textit{Colonization} in \texttt{D3}. 
Sometimes the confusion comes from changes in the writing style. For example \textit{arriver (to arrive)} means \textit{Arriving} in both \texttt{D1} and \texttt{D2}, but in \texttt{D3} it is used as a modal verb (\textit{arriver \`{a}} meaning \textit{to be able to}) triggering no frame.Under these circumstances, a model trained on a single domain  underestimates the complexity of the frame identification task, mapping the LU to the frame without further analysis of its sense.

When we apply $biGRU+AC$, the gains observed on Frame Identification are not constant across LUs. To analyze the impact of the adversarial approach, we compare for each LU the distribution across clusters of the sentences containing the given LU. This is done separately for \texttt{D1} and \texttt{D3} (for \texttt{D3}, sentences are projected into the clusters by choosing the less distant cluster centroid). In table \ref{tab:LU_FrameID}, we present the LUs for which the cluster distribution on \texttt{D1} and \texttt{D3} are the most dissimilar. These are also the LUs that are the most positively affected by the adversarial strategy. 

This means that whenever a LU has similar distribution of context words across the different domains this context information is already exploited by the system to perform frame disambiguation. On the other hand, when the context words of a LU depend on the domain, the model can take advantage of adversarial learning to build a higher level representation that abstracts as much as possible from the lexical variations of the words surrounding the LU.  



\subsubsection{Argument identification}

In this section, we focus on the Frame Argument (or FE for Frame Element) Identification level, and propose contrastive experiments following the complexity factors analysis proposed by \cite{marzinotto:hal-01731385}. In this study, the authors have shown that FE identification is performing better for verbal LU triggers than for nominal ones, and for triggers that are at the root of the syntactic dependency tree. We want to see here how these complexity factors are affected by the adversarial learning strategy. Additionally, we want to see if the system behaves equivalently over core and non-core Frame Elements.
Actually, in the usual evaluation Framework of the Framenet 1.5 shared task, non-core FEs are assigned a 0.5 weight for the F-measure computation, reducing the impact of errors on non-core FEs. In this paper, all FEs are rated equally, but here we separate them to observe their behaviour. As we can see in table \ref{tab:ArgIDperf}, adversarial training consistently improves the FE identification results, in all conditions. Moreover, bigger gains are observed for the \textit{difficult} conditions. 
  
\begin{table}
\centering
\small
\begin{tabular}{lcc}
\hline
\texttt{D3}   & $biGRU$ & $biGRU+AC$ \\
\hline
overall     & 50.2 & 51.7 (+3\%)\\ 
\hline
core FE     & 56.5 & 57.0 (+0.9\%)\\ 
non-core FE & 48.9 & 50.4 (+3.1\%)\\ 
\hline
verbal trigger    & 53.4 & 54.9 (+2.8\%)\\ 
nominal trigger   & 34.6 & 39.0 (+12.7\%)\\ 
\hline
root trigger     & 59.5 & 61.3 (+3.0\%)\\ 
non-root trigger & 45.4 & 47.2 (+4.0\%)\\
\hline
\end{tabular}
\normalsize
\caption{Frame Element Identification results according to complexity factors on \texttt{D3} (Fmax)}
\label{tab:ArgIDperf}
\end{table}

\section{Generalization to PropBank Parsing}


We further show that this adversarial learning technique can be used on other semantic frameworks such as Propbank. In PropBank Semantic Role Labeling, CoNLL-2005 uses Wall Street Journal (WSJ) for training and two test corpora. The in-domain (ID) test set is derived from WSJ and the out-of-domain (OOD) test set contains 'general fiction' from the Brown corpus. In published works, there is always an important gap in performances between ID and OOD. Several studies have tried to develop models with better generalization capacities \cite{yang2015domain}, \cite{fitzgerald2015semantic}. 
In recent works, PropBank SRL systems have evolved and span classifier approaches have been replaced by current state of the art sequence tagging models that use recurrent neural networks \cite{he2017deep} and neural attention \cite{DBLP:journals/corr/abs-1712-01586,strubell2018linguistically}. 
However, these parsers still suffer performances drops of up to 12 points in F-measure on OOD with respect to ID. 
For this experiment we have applied the same adversarial approach over a state-of-the-art Propbank parser \cite{he2017deep} using a single straight classifier model. As there is no explicit domain partitions in the training corpus, we apply our inferred domain adversarial task approach, running the clustering algorithm with 5 clusters. We were not able to reproduce the same results as the one published in the paper, we hence provide the results obtained running the downloaded system in our lab. Similarly to the previous FrameNet parsing model, we vary a threshold on the output probabilities of Semantic Roles in order to optimize the F-measure and we provide F-max values, computed using the official evaluation script. 
We observe in table \ref{tab:SRLperf} that the adversarial approach outperforms the original system on both the ID and OOD tests sets.

\begin{table}
\centering
\small
\begin{tabular}{lcc}
\hline
 \multicolumn{1}{p{1.1cm}}{}  & ID WSJ& OOD BROWN\\
\hline
\multicolumn{1}{l}{\cite{he2017deep}} & 82.4 & 71.7 \\ 
\multicolumn{1}{l}{\cite{he2017deep}+ ${AC}$} & 83.0 &  72.3 \\ 
\hline
\end{tabular}
\normalsize
\caption{SRL performance (Fmax) on CoNLL-2005, based on \cite{he2017deep} model}
\label{tab:SRLperf}
\end{table}


\section{Conclusion}
We have presented a study on improving the robustness of a frame semantic parser using adversarial learning. Results obtained on a multi-domain publicly available benchmark, called \textit{CALOR-Frame}, showed that domain adversarial training can effectively be used to improve the generalization capacities of the tagging models, even without prior information about domains in the training corpus. We showed that our technique can be applied to other semantic models, by implementing it into a state-of-the-art PropBank parser and showing some consistent gains. This positive result suggests that our approach could apply successfully to more NLP tasks.

\bibliography{naaclhlt2019}
\bibliographystyle{acl_natbib}

\appendix

\end{document}